\documentclass[sigconf]{acmart}
\AtBeginDocument{%
  }

\copyrightyear{2025}
\acmYear{2025}
\setcopyright{acmlicensed}\acmConference[WWW Companion '25]{Companion
Proceedings of the ACM Web Conference 2025}{April 28-May 2, 2025}{Sydney,
NSW, Australia}
\acmBooktitle{Companion Proceedings of the ACM Web Conference 2025 (WWW
Companion '25), April 28-May 2, 2025, Sydney, NSW, Australia}
\acmDOI{10.1145/XXXXXX.XXXXXX}
\acmISBN{979-8-4007-1331-6/25/04}

\begin{document}

\title{FMDLlama: Financial Misinformation Detection Based on Large Language Models}


\author{Zhiwei Liu}
\affiliation{%
  \institution{The University of Manchester}
  \city{Manchester}
  \country{United Kingdom}
}
\email{zhiwei.liu@manchester.ac.uk}

\author{Xin Zhang}
\affiliation{%
  \institution{The University of Manchester}
  \city{Manchester}
  \country{United Kingdom}
}
\email{xin.zhang-41@postgrad.manchester.ac.uk}

\author{Kailai Yang}
\affiliation{%
  \institution{The University of Manchester}
  \city{Manchester}
  \country{United Kingdom}
}
\email{kailai.yang@manchester.ac.uk}

\author{Qianqian Xie}
\authornote{Corresponding author}
\affiliation{%
  \institution{The Fin AI}
  \city{Singapore}
  \country{Singapore}
}
\email{xqq.sincere@gmail.com}

\author{Jimin Huang}
\affiliation{%
  \institution{The Fin AI}
  \city{Singapore}
  \country{Singapore}
}
\email{jimin.huang@thefin.ai}

\author{Sophia Ananiadou}
\affiliation{%
  \institution{The University of Manchester}
  \city{Manchester}
  \country{UK}
}
\affiliation{%
  \institution{Archimedes RC}
  \city{Athena}
  \country{Greece}
}
\email{sophia.ananiadou@manchester.ac.uk}

\renewcommand{\shortauthors}{Zhiwei Liu et al.}

\begin{abstract}
  The emergence of social media has made the spread of misinformation easier. In the financial domain, the accuracy of information is crucial for various aspects of financial market, which has made financial misinformation detection (FMD) an urgent problem that needs to be addressed. Large language models (LLMs) have demonstrated outstanding performance in various fields. However, current studies mostly rely on traditional methods and have not explored the application of LLMs in the field of FMD. The main reason is the lack of FMD instruction tuning datasets and evaluation benchmarks.  In this paper, we propose FMDLlama, the first open-sourced instruction-following LLMs for FMD task based on fine-tuning Llama3.1 with instruction data, the first multi-task FMD instruction dataset (FMDID) to support LLM instruction tuning, and a comprehensive FMD evaluation benchmark (FMD-B) with classification and explanation generation tasks to test the FMD ability of LLMs.  We compare our models with a variety of LLMs on FMD-B, where our model outperforms other open-sourced LLMs as well as OpenAI's products. This project is available at https://github.com/lzw108/FMD.
\end{abstract}



\maketitle

\section{Introduction}
In the financial sector, the accuracy of information is crucial for the integrity of decisions, market operation, risk management, compliance, and trust establishment \cite{rangapur2023investigating}. However, the proliferation of digital media has escalated the spread of financial misinformation \cite{chung2023theory}. Such misinformation, including deceptive investment propositions and biased news articles, can manipulate market prices and influence economic sentiment, presenting substantial risks \cite{kogan2020fake}. Furthermore, manually checking financial misinformation consumes a large amount of time and manpower \cite{kamal2023financial}. Therefore, the automatic identification of financial misinformation is an urgent priority for the normal operation of financial activities, yet there is currently limited exploration by researchers in this area.

Recently, large Language Models (LLMs) with large parameters have been explored as a new approach to address various issues in the financial domain \cite{shah2022flue}, yielding promising results. However, most studies concentrate on applying traditional deep learning methods such as CNNs, LSTMs, or pre-trained language models (PLMs) with fewer parameters like BERT or RoBERTa to detect false information in the financial domain \cite{kamal2023financial,chung2023theory,mohankumar2023financial}. \citet{rangapur2023finfact} evaluate a few LLMs on the FMD task. There are currently no LLMs specifically designed to detect financial misinformation. The main reason is the lack of data available for instruction-tuning LLMs.

To address the above issues, we construct the first instruction-tuning datasets for financial misinformation detection (FMDID) to support LLMs fine-tuning, including classification and explanation tasks. We subsequently propose the first open-sourced financial misinformation detection LLMs (FMDLlama) based on fine-tuning Llama3.1-8b-instruct with FMDID. To evaluate the financial misinformation verification ability of LLMs, we also build a benchmark for the detection of financial misinformation (FMD-B). The results on FMD-B show that FMDLlama achieves state-of-the-art (SOTA) performance among other open-sourced LLMs, as well as the closed-sourced OpenAI's products. 

Our main contributions are as follows:

(1) We construct the FMDID, the first multi-task financial misinformation
instruction-tuning dataset. 

(2) We develop FMDLlama, the first open-sourced financial misinformation detection LLMs that are specialized for diverse financial misinformation detection tasks.

(3) We build FMD-B, the first benchmark to evaluate the verification ability of financial misinformation of LLMs. The results on FMD-B demonstrate that our model overtakes other open-sourced LLMs and GPT-3.5-turbo, GPT-4o, and GPT-4o-mini.

\section{Related Work}

\subsection{Financial Misinformation Detection}
There are few studies detecting financial misinformation. Most of them are based on traditional deep learning methods or PLMs. \citet{kamal2023financial} introduce a framework for FMD task based on RoBERTa and multi-channel networks (CNNs, BiGRU, and attention layers). \citet{chung2023theory} apply multiple LSTMs to learn dynamic and hidden patterns to support financial disinformation detection.  \citet{mohankumar2023financial} adopt two cross-joint networks to build contextual sequential representation, which is produced by the combination of context-aware linguistic and financial embeddings, to detect fake news. \citet{rangapur2023finfact} propose one dataset for financial fact-checking and explanation generation, and evaluate the ability of several LLMs (e.g. GPT-4, Claude3, Mistral) on this dataset. However, there is currently no open-sourced LLM specifically designed for the detection of financial misinformation.

\subsection{Open Sourced Large Language Models}

Significant research efforts have focused on creating open-sourced LLMs as alternatives to the closed-sourced models (e.g. ChatGPT, GPT-4), which aims to facilitate more accessible research into enhancing and applying LLMs.  Well-known series of open-sourced, general-purpose language models include Llama series (e.g. llama2, llama3, llama3.1) \cite{touvron2023llama2,dubey2024llama}, Vicuna-7B-v1.5\footnote{https://huggingface.co/lmsys/vicuna-13b-v1.5}, Gemma \cite{team2024gemma1}, Mistral \cite{jiang2023mistral4}. There are also many open-sourced LLMs for specific domains, including FinMA \cite{xie2023pixiu2} for finance, MentalLLaMA \cite{yang2023mentalllama3} for mental health, ExTES-LLaMA \cite{zheng2023building4} for emotional support chatbots, and EmoLLMs \cite{liu2024emollms} for sentiment analysis and ConspEmoLLM \cite{liu2024conspemollm} for conspiracy detection. In this work, we extend the inventory of domain-specific LLMs, by developing the first open-sourced LLM for multitask financial misinformation detection.

\section{Methods}

\subsection{Task Formalization}

We approach financial misinformation detection as a generative task, applying a generative model as a foundation. This generative model is an autoregressive language model $P_{\phi}(y|x)$, parameterized using pre-trained weights $\phi$. It has the ability to simultaneously handle multiple financial misinformation detection tasks, i.e.,  misinformation detection, and explanation generation. Each task ($t$) is represented as a set of context-target pairs: $D_t={(q_i^t,r_i^t)}_{i={1,2,...N_t}}$, where the context $q$ is a token sequence containing the task description, input text, and query, and $r$ is a further token sequence containing the answer to the query. The model is optimized based on the merged dataset, which combines all task datasets, with the aim of maximizing the objective of conditional language modeling to improve prediction and generation performance.

\subsection{Construction of Instruction Tuning Dataset}

\subsubsection{Raw Data}

We build our instruction tuning dataset using two existing datasets.

\textbf{FinFact } FinFact \cite{rangapur2023finfact} is a comprehensive collection of financial claims categorized into areas like Income, Finance, Economy, Budget, Taxes, and Debt. The claim label categorizes claims as 'True', 'False', and 'NEI (Not Enough Information)'. It is meticulously crafted to reflect the complexity of financial narratives, including contextual details, supporting evidence links, and visual elements like image links and captions for each claim. A distinctive feature of this dataset is that it provides explanations for why each claim is deemed true or false, enriching the dataset's utility for training LLMs in not just detecting misinformation but also articulating reasoned explanations for their assessments. 

\textbf{FinGuard } Financial Truth Guard dataset\footnote{https://github.com/carlos-gmartin/Financial-Truth-Guard} is used for analyzing news articles and predicting whether they contain misleading or fake information related to the financial markets. 

\begin{table}[htb]
\small
\caption{\label{tab:datastatistics}
Dataset statistics. \textit{Raw} denotes the raw data from FinFact and FinGuard. \textit{Instruction} denotes the converted instruction data based on \textit{Raw}.}
\begin{tabular}{lcccccc}
\hline
Data     & \multicolumn{3}{c}{Raw} & \multicolumn{3}{c}{Instruction} \\
         & Train   & Val   & Test  & Train      & Val     & Test     \\ \hline
FinFact  & 1562    & 391   & 1304  & 1562       & 391     & 1304     \\
FinGuard & 2900    & 600   & 1500  & 2900       & 600     & 1500     \\ \hline
\end{tabular}
\end{table}

\subsubsection{Construction of the FMD instruction tuning dataset (FMDID) and FMD benchmark (FMD-B)}

\begin{figure*}[htb]
\centering
\includegraphics[width=1.7\columnwidth]{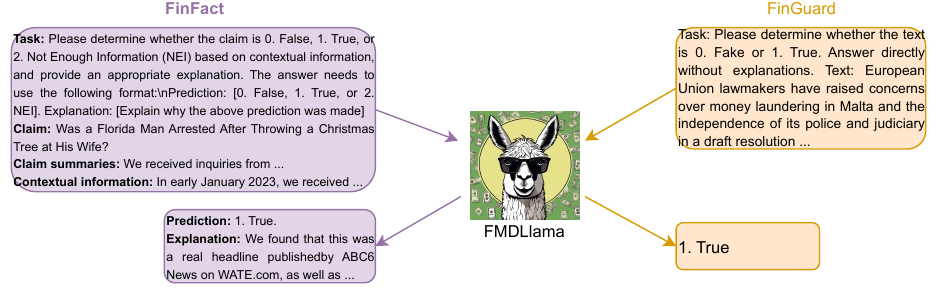}
\caption{An overview of multi-task instruction tuning of FMDLlama}
\label{fig:FMDLLMs}
\end{figure*}

We use the raw datasets as the basis to build the instruction dataset. For FinFact, We split the original data into train, validation, and test sets and remove the data without evidence (i.e. explanation). For FinGuard, we separately extracted 2500 data points from both real and fake data and divided them into the train, validation, and test sets. The dataset statistics are presented in Table \ref{tab:datastatistics}. We construct instruction-tuning data for each task based on the template in Table \ref{tab:taskprompt}. For FinFact, \textit{[raw claim]}, \textit{[raw summaries]}, and \textit{[raw contextual]} are from the raw data. The format of LLMs' response will be Prediction: \textit{[Lable]}. Explanation: \textit{[Explanations]}. \textit{[Lable]} is one of \textit{[0. False, 1. True, or 2. NEI]}. \textit{[Explanations]} is the reason why the LLM makes the \textit{[Lable]} decision. For FinFact, \textit{[Explanations]} is from the raw data. Figure \ref{fig:FMDLLMs} presents examples used to fine-tune the LLM.

\begin{table*}[htb]
\footnotesize
\caption{\label{tab:taskprompt} Instructions used for each task.}
\begin{tabular}{ll}
\hline
Task     & Instruction Template                                                                                                                                                                                                                                                                                                                                                                                                                                                        \\ \hline
FinFact  & \begin{tabular}[c]{@{}l@{}}Task: Please determine whether the claim is 0. False, 1. True, or 2. Not Enough Information (NEI) \\ based on contextual information, and provide an appropriate explanation.\\ The answer needs to use the following format:\\ Prediction: [0. False, 1. True, or 2. NEI]\\ Explanation: [Explain why the above prediction was made]\\ Claim: [raw claim]. Claim summaries: [raw summaries]. Contextual information: [raw contextual]\end{tabular} \\
FinGuard & \begin{tabular}[c]{@{}l@{}}Task: Please determine whether the text is 0. Fake or 1. True. Answer directly without explanations. \\ Text: [input text]\end{tabular}                                                                                                                                                                                                                                                                                                          \\ \hline
\end{tabular}
\end{table*}

After constructing the instruction data, we collect the train and validation data as instruction-tuning data (FMDID) and test data as the Financial Misinformation Detection benchmark (FMD-B), which are used to fine-tune the LLMs and evaluate the ability of LLMs in the Financial Misinformation Detection domain respectively.

\subsection{FMDLLMs}

We built FMDLlama2 and FMDLlama3 by fine-tuning Llama2-chat-7b \cite{touvron2023llama2} and Llama-3.1-8b-instruct\footnote{https://huggingface.co/meta-llama/Meta-Llama-3.1-8B-Instruct} using the FMDID dataset. The models are trained based on the AdamW optimizer \cite{loshchilov2017decoupled4} for three epochs, using DeepSpeed \cite{rasley2020deepspeed1} to reduce memory usage. We set the batch size to 128. The initial learning rate is set to 1e-6 with a warm-up ratio of 5\%. All models are trained on two Nvidia Tesla A100 GPUs, each with 80GB of memory. Figure \ref{fig:FMDLLMs} provides an overview of multi-task instruction tuning of FMDLlama for diverse financial misinformation detection tasks.

\section{Experiments}

\begin{table*}[!t]
\footnotesize
\caption{\label{tab:results} Results on FMD-B. R1, R2, RL denote  ROUGE (1, 2, and L) respectively. Bold indicates the best performance. Underline indicates the second-best performance.}
\begin{tabular}{lcccccccccccc}
\hline
 & \multicolumn{8}{c}{FinFact}                                                                                                                   & \multicolumn{4}{c}{FinGuard}                                          \\
                        & \multicolumn{4}{c}{Classification}                                    & \multicolumn{4}{c}{Explanation}                                       & \multicolumn{4}{c}{Classification}                                    \\
Methods                 & ACC             & PRE             & REC             & F1              & Rouge1          & Rouge2          & Rouge3          & BERTscore       & ACC             & PRE             & REC             & F1              \\ \hline
BERT                    & 0.6221          & 0.5550          & 0.5153          & 0.4836          & -               & -               & -               & -               & 0.9845          & 0.9845          & 0.9845          & 0.9845          \\
RoBERTa                 & 0.6822          & 0.6373          & 0.5823          & 0.5661          & -               & -               & -               & -               & \textbf{0.9961} & \textbf{0.9962} & \textbf{0.9961} & \textbf{0.9961} \\ \hline
Gemma-2b-instruct               & 0.0782          & 0.3219          & 0.0853          & 0.0866          & 0.0295          & 0.0077          & 0.0208          & 0.0539          & 0.1047          & 0.2564          & 0.0513          & 0.0692          \\
Vicuna-7b                & 0.3758          & 0.4697          & 0.3406          & 0.2568          & 0.2712          & 0.1012          & 0.1677          & 0.5429          & 0.4167          & 0.3168          & 0.2746          & 0.2806          \\
Vicuna-13b               & 0.2991          & 0.3399          & 0.2748          & 0.2281          & 0.2593          & 0.0884          & 0.1568          & 0.5310          & 0.3707          & 0.2701          & 0.1844          & 0.2184          \\
Mistral-7b-instruct              & 0.6097          & 0.4857          & 0.4747          & 0.4476          & 0.2724          & 0.0859          & 0.1574          & 0.5492          & 0.5887          & 0.5905          & 0.5860          & 0.5823          \\
Llama2-chat-7b               & 0.3198          & 0.2436          & 0.1893          & 0.1902          & 0.1557          & 0.0495          & 0.0950          & 0.3443          & 0.3100          & 0.3735          & 0.2024          & 0.2079          \\
Llama2-chat-13b              & 0.1933          & 0.3528          & 0.1716          & 0.1871          & 0.0645          & 0.0181          & 0.0399          & 0.1538          & 0.3053          & 0.3363          & 0.2011          & 0.2354          \\
Llama3.1-8b-instruct             & 0.6449          & 0.6209          & 0.5652          & 0.5383          & 0.2111          & 0.0823          & 0.1394          & 0.5449          & 0.5600          & 0.6541          & 0.5502          & 0.4643          \\
GPT-3.5-turbo                 & \underline{0.7270}    & 0.6635          & \underline{0.6433}    & \underline{0.6380}    & 0.2639          & 0.1005          & 0.1645          & 0.5642          & 0.6800          & 0.4947          & 0.4492          & 0.4359          \\
GPT-4o-mini             & 0.7132          & \textbf{0.7185} & 0.6368          & 0.6296          & 0.2682          & 0.0818          & 0.1559          & 0.5582          & 0.6867          & 0.7846          & 0.6795          & 0.6519          \\
GPT-4o                  & 0.6902          & 0.6596          & 0.6271          & 0.6283          & \underline{0.2855}    & 0.0956          & 0.1663          & \underline{0.5668}    & 0.7720          & 0.5541          & 0.5112          & 0.5062          \\ \hline
FMDLlama2               & 0.6986          & 0.4820          & 0.4522          & 0.4534          & 0.2443          & \underline{0.1670}    & \underline{0.2011}    & 0.5214          & 0.9433          & 0.6440          & 0.6283          & 0.6354          \\
FMDLlama3             & \textbf{0.7362} & \underline{0.6733}    & \textbf{0.6700} & \textbf{0.6667} & \textbf{0.4524} & \textbf{0.3498} & \textbf{0.3773} & \textbf{0.6756} & \underline{0.9947}    & \underline{0.9947}    & \underline{0.9947}    & \underline{0.9947}    \\ \hline
\end{tabular}
\end{table*}

\begin{table}[]
\begin{tabular}{lcccccccccccc}

\end{tabular}
\end{table}

\subsection{Baseline Models}

\textbf{PLMs:} Financial misinformation detection is typically regarded as a classification task. For our baseline models, we select commonly used PLMs, which can only be fine-tuned for individual tasks, i.e., the general language BERT \cite{devlin2018bert6} and RoBERTa \cite{liu2019roberta7}. We treat FinFact as a 3-way classification task, and FinGuard as a binary classification task, using cross-entropy loss for training. 

\textbf{LLMs:} LLMs have been proven to be capable of solving numerous tasks. We apply zero-shot prompting on the instruction dataset to the following open-sourced LLMs:  Llama2-chat-(7b,13b) \cite{touvron2023llama2}, Llama3.1-8b-instruct\footnote{https://www.llama.com/}, Gemma-2b-instruct \cite{team2024gemma1}, Mistral-7b-instruct \cite{jiang2023mistral4} and Vicuna-(7b,13b)-v1.5\footnote{https://huggingface.co/lmsys/vicuna-13b-v1.5}. We also utilize zero-shot prompting with the proprietary LLMs GPT-3.5-turbo, GPT-4o, and GPT-4o-mini \cite{achiam2023gpt}.

\subsection{Evaluation Methods}

We use metrics such as Accuracy, Precision, Recall, and Macro-F1 for misinformation detection (classification) evaluation and ROUGE (1, 2, and L) \cite{lin2004rouge}, BERTScore \cite{zhang2019bertscore} for explanation evaluation. 


\subsection{Results}

Table \ref{tab:results} presents the results on FMD-B. From the table, we can see FMDLlama3 achieves SOTA results among all other open-sourced LLMs as well as the close-source OpenAI series. Although BERT and RoBERTa were fine-tuned on each classification task separately and have similar results with FMDLlama3 on simple FinGuard dataset, their performance is lower than FMDLlama3 on the complex dataset FinFact. A possible reason is that it is challenging for the PLMs with less parameters to understand long and complex textual content. For LLMs without fine-tuning, Mistral-7b-instruct, Llama3.1-8b-instruct, and OpenAI series\footnote{From the table, we can see that GPT-4o and GPT-4o-mini are not as effective as GPT-3.5-turbo in multi-class tasks (i.e. FinFact classification). However, their explanation generation's Rouge-1 and BERTscore metrics and the score in the simple classification task (i.e. FinGuard) have improved compared to GPT-3.5-turbo.} perform well. This is because most of the data points in FMD-B are long texts. Mistral-7b-instruct, Llama3.1-8b-instruct and OpenAI series allow longer input lengths, and have a better understanding of corresponding long texts. By comparing the results of FMDLlama2 with Llama2-chat-7b and FMDLlama3 with Llama3.1-8b-instruct, we can recognize the effectiveness of instruction-tuning strategies. Instruction-tuning strategies in specific domains allow LLMs to focus more on those domains, leading to better performance. Overall, the results from the table indicate that in the field of financial misinformation detection, our open-sourced model FMDLlama3 with 8b parameters has surpassed the closed-sourced OpenAI series.


\section{Conclusion}

In this paper, we propose FMDLlama, the first LLM for financial misinformation detection (FMD). We also construct a multi-task FMD instruction dataset (FMDID) and a FMD evaluation benchmark (FMD-B). We conduct a comprehensive analysis of the performance of FMDLlama, as well as a variety of LLMs on the FMD-B benchmark. The results indicate that FMDLlama performs exceptionally well in FMD tasks, achieving SOTA compared to the other open-sourced LLMs as well as OpenAI's products.

In the future, we aim to augment the FMDID and FMD-B datasets with further FMD datasets, including data from multiple platforms, sources, domains and languages, which can help further improve the FMDLlama and evaluate the FMD ability of LLMs more comprehensively.

\section{Limitations}

The potential limitations of our work may be summarized as follows:

(1) Due to restricted computational resources, we only carried out instruction-tuning/evaluation of financial misinformation detection tasks using 2b/7b/8b/13b LLMs. As such, we have not considered the impact of using larger models on the FMD tasks.

(2) Due to the limited availability of publicly accessible datasets on financial misinformation, we constructed instruction-tuning datasets and benchmarks for financial misinformation detection based solely on two datasets, two kinds of tasks (i.e. classification, explanation generation).

\begin{acks}
The FMDLlama illustration in Figure \ref{fig:FMDLLMs} was generated using PIXLR\footnote{https://pixlr.com/image-generator/}. This work is supported by the computational shared facility at the University of Manchester and the scholar award from the Department of Computer Science at the University of Manchester. This work is supported by the project JPNP20006 from New Energy and Industrial Technology Development Organization (NEDO) and the Manchester-Melbourne-Toronto Research Funding. This work has been partially supported by project MIS 5154714 of the National Recovery and Resilience Plan Greece 2.0 funded by the European Union under the NextGeneration EU Program.
\end{acks}

\bibliographystyle{ACM-Reference-Format}
\balance
\bibliography{sample-base}


\begin{thebibliography}{23}


\ifx \showCODEN    \undefined \def \showCODEN     #1{\unskip}     \fi
\ifx \showDOI      \undefined \def \showDOI       #1{#1}\fi
\ifx \showISBNx    \undefined \def \showISBNx     #1{\unskip}     \fi
\ifx \showISBNxiii \undefined \def \showISBNxiii  #1{\unskip}     \fi
\ifx \showISSN     \undefined \def \showISSN      #1{\unskip}     \fi
\ifx \showLCCN     \undefined \def \showLCCN      #1{\unskip}     \fi
\ifx \shownote     \undefined \def \shownote      #1{#1}          \fi
\ifx \showarticletitle \undefined \def \showarticletitle #1{#1}   \fi
\ifx \showURL      \undefined \def \showURL       {\relax}        \fi
\providecommand\bibfield[2]{#2}
\providecommand\bibinfo[2]{#2}
\providecommand\natexlab[1]{#1}
\providecommand\showeprint[2][]{arXiv:#2}

\bibitem[Achiam et~al\mbox{.}(2023)]%
        {achiam2023gpt}
\bibfield{author}{\bibinfo{person}{Josh Achiam}, \bibinfo{person}{Steven Adler}, \bibinfo{person}{Sandhini Agarwal}, \bibinfo{person}{Lama Ahmad}, \bibinfo{person}{Ilge Akkaya}, \bibinfo{person}{Florencia~Leoni Aleman}, \bibinfo{person}{Diogo Almeida}, \bibinfo{person}{Janko Altenschmidt}, \bibinfo{person}{Sam Altman}, \bibinfo{person}{Shyamal Anadkat}, {et~al\mbox{.}}} \bibinfo{year}{2023}\natexlab{}.
\newblock \showarticletitle{Gpt-4 technical report}.
\newblock \bibinfo{journal}{\emph{arXiv preprint arXiv:2303.08774}} (\bibinfo{year}{2023}).
\newblock


\bibitem[Chung et~al\mbox{.}(2023)]%
        {chung2023theory}
\bibfield{author}{\bibinfo{person}{Wingyan Chung}, \bibinfo{person}{Yinqiang Zhang}, {and} \bibinfo{person}{Jia Pan}.} \bibinfo{year}{2023}\natexlab{}.
\newblock \showarticletitle{A theory-based deep-learning approach to detecting disinformation in financial social media}.
\newblock \bibinfo{journal}{\emph{Information Systems Frontiers}} \bibinfo{volume}{25}, \bibinfo{number}{2} (\bibinfo{year}{2023}), \bibinfo{pages}{473--492}.
\newblock


\bibitem[Devlin et~al\mbox{.}(2018)]%
        {devlin2018bert6}
\bibfield{author}{\bibinfo{person}{Jacob Devlin}, \bibinfo{person}{Ming-Wei Chang}, \bibinfo{person}{Kenton Lee}, {and} \bibinfo{person}{Kristina Toutanova}.} \bibinfo{year}{2018}\natexlab{}.
\newblock \showarticletitle{Bert: Pre-training of deep bidirectional transformers for language understanding}.
\newblock \bibinfo{journal}{\emph{arXiv preprint arXiv:1810.04805}} (\bibinfo{year}{2018}).
\newblock


\bibitem[Dubey et~al\mbox{.}(2024)]%
        {dubey2024llama}
\bibfield{author}{\bibinfo{person}{Abhimanyu Dubey}, \bibinfo{person}{Abhinav Jauhri}, \bibinfo{person}{Abhinav Pandey}, \bibinfo{person}{Abhishek Kadian}, \bibinfo{person}{Ahmad Al-Dahle}, \bibinfo{person}{Aiesha Letman}, \bibinfo{person}{Akhil Mathur}, \bibinfo{person}{Alan Schelten}, \bibinfo{person}{Amy Yang}, \bibinfo{person}{Angela Fan}, {et~al\mbox{.}}} \bibinfo{year}{2024}\natexlab{}.
\newblock \showarticletitle{The llama 3 herd of models}.
\newblock \bibinfo{journal}{\emph{arXiv preprint arXiv:2407.21783}} (\bibinfo{year}{2024}).
\newblock


\bibitem[Jiang et~al\mbox{.}(2023)]%
        {jiang2023mistral4}
\bibfield{author}{\bibinfo{person}{Albert~Q Jiang}, \bibinfo{person}{Alexandre Sablayrolles}, \bibinfo{person}{Arthur Mensch}, \bibinfo{person}{Chris Bamford}, \bibinfo{person}{Devendra~Singh Chaplot}, \bibinfo{person}{Diego de~las Casas}, \bibinfo{person}{Florian Bressand}, \bibinfo{person}{Gianna Lengyel}, \bibinfo{person}{Guillaume Lample}, \bibinfo{person}{Lucile Saulnier}, {et~al\mbox{.}}} \bibinfo{year}{2023}\natexlab{}.
\newblock \showarticletitle{Mistral 7B}.
\newblock \bibinfo{journal}{\emph{arXiv preprint arXiv:2310.06825}} (\bibinfo{year}{2023}).
\newblock


\bibitem[Kamal et~al\mbox{.}(2023)]%
        {kamal2023financial}
\bibfield{author}{\bibinfo{person}{Ashraf Kamal}, \bibinfo{person}{Padmapriya Mohankumar}, {and} \bibinfo{person}{Vishal~Kumar Singh}.} \bibinfo{year}{2023}\natexlab{}.
\newblock \showarticletitle{Financial Misinformation Detection via RoBERTa and Multi-channel Networks}. In \bibinfo{booktitle}{\emph{International Conference on Pattern Recognition and Machine Intelligence}}. Springer, \bibinfo{pages}{646--653}.
\newblock


\bibitem[Kogan et~al\mbox{.}(2020)]%
        {kogan2020fake}
\bibfield{author}{\bibinfo{person}{Shimon Kogan}, \bibinfo{person}{Tobias~J Moskowitz}, {and} \bibinfo{person}{Marina Niessner}.} \bibinfo{year}{2020}\natexlab{}.
\newblock \bibinfo{booktitle}{\emph{Fake news in financial markets}}.
\newblock \bibinfo{publisher}{SSRN}.
\newblock


\bibitem[Lin(2004)]%
        {lin2004rouge}
\bibfield{author}{\bibinfo{person}{Chin-Yew Lin}.} \bibinfo{year}{2004}\natexlab{}.
\newblock \showarticletitle{Rouge: A package for automatic evaluation of summaries}. In \bibinfo{booktitle}{\emph{Text summarization branches out}}. \bibinfo{pages}{74--81}.
\newblock


\bibitem[Liu et~al\mbox{.}(2019)]%
        {liu2019roberta7}
\bibfield{author}{\bibinfo{person}{Yinhan Liu}, \bibinfo{person}{Myle Ott}, \bibinfo{person}{Naman Goyal}, \bibinfo{person}{Jingfei Du}, \bibinfo{person}{Mandar Joshi}, \bibinfo{person}{Danqi Chen}, \bibinfo{person}{Omer Levy}, \bibinfo{person}{Mike Lewis}, \bibinfo{person}{Luke Zettlemoyer}, {and} \bibinfo{person}{Veselin Stoyanov}.} \bibinfo{year}{2019}\natexlab{}.
\newblock \showarticletitle{Roberta: A robustly optimized bert pretraining approach}.
\newblock \bibinfo{journal}{\emph{arXiv preprint arXiv:1907.11692}} (\bibinfo{year}{2019}).
\newblock


\bibitem[Liu et~al\mbox{.}(2024a)]%
        {liu2024conspemollm}
\bibfield{author}{\bibinfo{person}{Zhiwei Liu}, \bibinfo{person}{Boyang Liu}, \bibinfo{person}{Paul Thompson}, \bibinfo{person}{Kailai Yang}, \bibinfo{person}{Raghav Jain}, {and} \bibinfo{person}{Sophia Ananiadou}.} \bibinfo{year}{2024}\natexlab{a}.
\newblock \showarticletitle{ConspEmoLLM: Conspiracy Theory Detection Using an Emotion-Based Large Language Model}.
\newblock \bibinfo{journal}{\emph{arXiv preprint arXiv:2403.06765}} (\bibinfo{year}{2024}).
\newblock


\bibitem[Liu et~al\mbox{.}(2024b)]%
        {liu2024emollms}
\bibfield{author}{\bibinfo{person}{Zhiwei Liu}, \bibinfo{person}{Kailai Yang}, \bibinfo{person}{Qianqian Xie}, \bibinfo{person}{Tianlin Zhang}, {and} \bibinfo{person}{Sophia Ananiadou}.} \bibinfo{year}{2024}\natexlab{b}.
\newblock \showarticletitle{Emollms: A series of emotional large language models and annotation tools for comprehensive affective analysis}. In \bibinfo{booktitle}{\emph{Proceedings of the 30th ACM SIGKDD Conference on Knowledge Discovery and Data Mining}}. \bibinfo{pages}{5487--5496}.
\newblock


\bibitem[Loshchilov and Hutter(2017)]%
        {loshchilov2017decoupled4}
\bibfield{author}{\bibinfo{person}{Ilya Loshchilov} {and} \bibinfo{person}{Frank Hutter}.} \bibinfo{year}{2017}\natexlab{}.
\newblock \showarticletitle{Decoupled weight decay regularization}.
\newblock \bibinfo{journal}{\emph{arXiv preprint arXiv:1711.05101}} (\bibinfo{year}{2017}).
\newblock


\bibitem[Mohankumar et~al\mbox{.}(2023)]%
        {mohankumar2023financial}
\bibfield{author}{\bibinfo{person}{Padmapriya Mohankumar}, \bibinfo{person}{Ashraf Kamal}, \bibinfo{person}{Vishal~Kumar Singh}, {and} \bibinfo{person}{Amrish Satish}.} \bibinfo{year}{2023}\natexlab{}.
\newblock \showarticletitle{Financial fake news detection via context-aware embedding and sequential representation using cross-joint networks}. In \bibinfo{booktitle}{\emph{2023 15th International Conference on COMmunication Systems \& NETworkS (COMSNETS)}}. IEEE, \bibinfo{pages}{780--784}.
\newblock


\bibitem[Rangapur et~al\mbox{.}(2023a)]%
        {rangapur2023finfact}
\bibfield{author}{\bibinfo{person}{Aman Rangapur}, \bibinfo{person}{Haoran Wang}, {and} \bibinfo{person}{Kai Shu}.} \bibinfo{year}{2023}\natexlab{a}.
\newblock \bibinfo{title}{Fin-Fact: A Benchmark Dataset for Multimodal Financial Fact Checking and Explanation Generation}.
\newblock
\newblock
\showeprint[arxiv]{2309.08793}~[cs.AI]


\bibitem[Rangapur et~al\mbox{.}(2023b)]%
        {rangapur2023investigating}
\bibfield{author}{\bibinfo{person}{Aman Rangapur}, \bibinfo{person}{Haoran Wang}, {and} \bibinfo{person}{Kai Shu}.} \bibinfo{year}{2023}\natexlab{b}.
\newblock \showarticletitle{Investigating online financial misinformation and its consequences: A computational perspective}.
\newblock \bibinfo{journal}{\emph{arXiv preprint arXiv:2309.12363}} (\bibinfo{year}{2023}).
\newblock


\bibitem[Rasley et~al\mbox{.}(2020)]%
        {rasley2020deepspeed1}
\bibfield{author}{\bibinfo{person}{Jeff Rasley}, \bibinfo{person}{Samyam Rajbhandari}, \bibinfo{person}{Olatunji Ruwase}, {and} \bibinfo{person}{Yuxiong He}.} \bibinfo{year}{2020}\natexlab{}.
\newblock \showarticletitle{Deepspeed: System optimizations enable training deep learning models with over 100 billion parameters}. In \bibinfo{booktitle}{\emph{Proceedings of the 26th ACM SIGKDD International Conference on Knowledge Discovery \& Data Mining}}. \bibinfo{pages}{3505--3506}.
\newblock


\bibitem[Shah et~al\mbox{.}(2022)]%
        {shah2022flue}
\bibfield{author}{\bibinfo{person}{Raj~Sanjay Shah}, \bibinfo{person}{Kunal Chawla}, \bibinfo{person}{Dheeraj Eidnani}, \bibinfo{person}{Agam Shah}, \bibinfo{person}{Wendi Du}, \bibinfo{person}{Sudheer Chava}, \bibinfo{person}{Natraj Raman}, \bibinfo{person}{Charese Smiley}, \bibinfo{person}{Jiaao Chen}, {and} \bibinfo{person}{Diyi Yang}.} \bibinfo{year}{2022}\natexlab{}.
\newblock \showarticletitle{When flue meets flang: Benchmarks and large pre-trained language model for financial domain}.
\newblock \bibinfo{journal}{\emph{arXiv preprint arXiv:2211.00083}} (\bibinfo{year}{2022}).
\newblock


\bibitem[Team et~al\mbox{.}(2024)]%
        {team2024gemma1}
\bibfield{author}{\bibinfo{person}{Gemma Team}, \bibinfo{person}{Thomas Mesnard}, \bibinfo{person}{Cassidy Hardin}, \bibinfo{person}{Robert Dadashi}, \bibinfo{person}{Surya Bhupatiraju}, \bibinfo{person}{Shreya Pathak}, \bibinfo{person}{Laurent Sifre}, \bibinfo{person}{Morgane Rivi{\`e}re}, \bibinfo{person}{Mihir~Sanjay Kale}, \bibinfo{person}{Juliette Love}, {et~al\mbox{.}}} \bibinfo{year}{2024}\natexlab{}.
\newblock \showarticletitle{Gemma: Open models based on gemini research and technology}.
\newblock \bibinfo{journal}{\emph{arXiv preprint arXiv:2403.08295}} (\bibinfo{year}{2024}).
\newblock


\bibitem[Touvron et~al\mbox{.}(2023)]%
        {touvron2023llama2}
\bibfield{author}{\bibinfo{person}{Hugo Touvron}, \bibinfo{person}{Louis Martin}, \bibinfo{person}{Kevin Stone}, \bibinfo{person}{Peter Albert}, \bibinfo{person}{Amjad Almahairi}, \bibinfo{person}{Yasmine Babaei}, \bibinfo{person}{Nikolay Bashlykov}, \bibinfo{person}{Soumya Batra}, \bibinfo{person}{Prajjwal Bhargava}, \bibinfo{person}{Shruti Bhosale}, {et~al\mbox{.}}} \bibinfo{year}{2023}\natexlab{}.
\newblock \showarticletitle{Llama 2: Open foundation and fine-tuned chat models}.
\newblock \bibinfo{journal}{\emph{arXiv preprint arXiv:2307.09288}} (\bibinfo{year}{2023}).
\newblock


\bibitem[Xie et~al\mbox{.}(2023)]%
        {xie2023pixiu2}
\bibfield{author}{\bibinfo{person}{Qianqian Xie}, \bibinfo{person}{Weiguang Han}, \bibinfo{person}{Xiao Zhang}, \bibinfo{person}{Yanzhao Lai}, \bibinfo{person}{Min Peng}, \bibinfo{person}{Alejandro Lopez-Lira}, {and} \bibinfo{person}{Jimin Huang}.} \bibinfo{year}{2023}\natexlab{}.
\newblock \showarticletitle{PIXIU: A Large Language Model, Instruction Data and Evaluation Benchmark for Finance}.
\newblock \bibinfo{journal}{\emph{arXiv preprint arXiv:2306.05443}} (\bibinfo{year}{2023}).
\newblock


\bibitem[Yang et~al\mbox{.}(2023)]%
        {yang2023mentalllama3}
\bibfield{author}{\bibinfo{person}{Kailai Yang}, \bibinfo{person}{Tianlin Zhang}, \bibinfo{person}{Ziyan Kuang}, \bibinfo{person}{Qianqian Xie}, {and} \bibinfo{person}{Sophia Ananiadou}.} \bibinfo{year}{2023}\natexlab{}.
\newblock \showarticletitle{Mentalllama: Interpretable mental health analysis on social media with large language models}.
\newblock \bibinfo{journal}{\emph{arXiv preprint arXiv:2309.13567}} (\bibinfo{year}{2023}).
\newblock


\bibitem[Zhang et~al\mbox{.}(2019)]%
        {zhang2019bertscore}
\bibfield{author}{\bibinfo{person}{Tianyi Zhang}, \bibinfo{person}{Varsha Kishore}, \bibinfo{person}{Felix Wu}, \bibinfo{person}{Kilian~Q Weinberger}, {and} \bibinfo{person}{Yoav Artzi}.} \bibinfo{year}{2019}\natexlab{}.
\newblock \showarticletitle{Bertscore: Evaluating text generation with bert}.
\newblock \bibinfo{journal}{\emph{arXiv preprint arXiv:1904.09675}} (\bibinfo{year}{2019}).
\newblock


\bibitem[Zheng et~al\mbox{.}(2023)]%
        {zheng2023building4}
\bibfield{author}{\bibinfo{person}{Zhonghua Zheng}, \bibinfo{person}{Lizi Liao}, \bibinfo{person}{Yang Deng}, {and} \bibinfo{person}{Liqiang Nie}.} \bibinfo{year}{2023}\natexlab{}.
\newblock \showarticletitle{Building emotional support chatbots in the era of llms}.
\newblock \bibinfo{journal}{\emph{arXiv preprint arXiv:2308.11584}} (\bibinfo{year}{2023}).
\newblock


\end{thebibliography}


\end{document}